\documentclass[10pt, a4paper, journal]{IEEEtran}
\IEEEoverridecommandlockouts
\usepackage{cite}
\usepackage{amsmath,amssymb,amsfonts}
\usepackage{algorithmic}
\usepackage{graphicx}
\usepackage{textcomp}
\usepackage{xcolor}
\usepackage{multirow}
\usepackage{bm}
\usepackage{orcidlink}
\def\BibTeX{{\rm B\kern-.05em{\sc i\kern-.025em b}\kern-.08em
    T\kern-.1667em\lower.7ex\hbox{E}\kern-.125emX}}
\begin{document}

\title{Sparse Low-Ranked Self-Attention Transformer for Remaining Useful Lifetime Prediction of Optical Fiber Amplifiers
\thanks{This work is supported by the SNS Joint Undertaken under grant agreement No. 101096120 (SEASON). Responsibility for the content of this publication is with the authors.}
}

\author{\IEEEauthorblockN{Dominic Schneider\textsuperscript{\normalsize\orcidlink{0009-0002-2612-2068}}, Lutz Rapp\textsuperscript{\normalsize\orcidlink{0000-0002-3410-4006}}}\\
\IEEEauthorblockA{\textit{Advanced Technology}, \textit{Adtran}, 98617 Meiningen, Germany \\
dominic.schneider@adtran.com}
}

\maketitle

\begin{abstract}
   Optical fiber amplifiers are key elements in present optical networks. Failures of these components result in high financial loss of income of the network operator as the communication traffic over an affected link is interrupted. Applying Remaining useful lifetime (RUL) prediction in the context of Predictive Maintenance (PdM) to optical fiber amplifiers to predict upcoming system failures at an early stage, so that network outages can be minimized through planning of targeted maintenance actions, ensures reliability and safety. Optical fiber amplifier are complex systems, that work under various operating conditions, which makes correct forecasting a difficult task. Increased monitoring capabilities of systems results in datasets that facilitate the application of data-driven RUL prediction methods. Deep learning models in particular have shown good performance, but generalization based on comparatively small datasets for RUL prediction is difficult. In this paper, we propose Sparse Low-ranked self-Attention Transformer (SLAT) as a novel RUL prediction method. SLAT is based on an encoder-decoder architecture, wherein two parallel working encoders extract features for sensors and time steps. By utilizing the self-attention mechanism, long-term dependencies can be learned from long sequences. The implementation of sparsity in the attention matrix and a low-rank parametrization reduce overfitting and increase generalization. Experimental application to optical fiber amplifiers exemplified on EDFA, as well as a reference dataset from turbofan engines, shows that SLAT outperforms the state-of-the-art methods.
\end{abstract}


\section{INTRODUCTION}
\IEEEPARstart{P}{redictive} maintenance (PdM) is an essential part of today's industry 4.0 and relies on condition monitoring, remaining useful lifetime (RUL) prognosis and fault case (FC) diagnosis of a system and its components in real-time  \cite{levitt2003complete}. The primary strategy is to plan mainenance actions as soon as a component shows atypical behavior that may lead to a FC. These faulty behaviors comprise degradation, which result in slightly decreased performance or complete failures of the system \cite{tinga2019physical}. Besides anomaly detection \cite{schneider2024anomaly} and FC diagnosis, a key component in modern PdM systems is represented by prognosing the RUL \cite{lughofer2019prologue}. Therefore, the condition of the system is monitored at specified inspection intervals from begin of life (BOL). This condition-based monitoring (CBM) requires the measurement of the systems parameters to reflect the health-state of the system in real-time \cite{cui2019research}. Based on CBM the RUL of the real-time operational state can be predicted to ensure maintenance decisions can be made accordingly.

Optical fiber amplifiers are key components in current long-haul optical fiber transmission networks. Interruptions of an optical transmission link, caused by system failure of an optical fiber amplifier, result in cost-intensive loss of transmission capacity. Applying PdM to predict the RUL of optical fiber amplifier and thus enabling the planning of targeted maintenance actions reduce network downtimes and increases the network resilience.

RUL prediction can be grouped into model-based and data-driven based approaches. Model-based prediction implies fundamental knowledge about the degradation processes at component level \cite{lei2018machinery}. With the increasing complexity in modern systems and non-linear dependencies of components and the system, the data-driven approach becomes more popular as accurate modeling using the traditional approach is difficult to realize.

The data-driven approach maps features of the system to a scalar value RUL utilizing an arbitrary model \cite{niu2010intelligent}. These features can contain sensor data as well as operating conditions. Furthermore, this approach requires no a priori knowledge about the physical mechanism that causes the degradation behavior of individual components and their interplay in the system. Various machine learning techniques have been applied to this task, whereas neural network based methods like multi-layer perceptron \cite{huang2007residual}, artificial neural networks \cite{tian2012artificial}, and fuzzy neural networks \cite{wang2004prognosis} showed promising results. However, these methods make elaborate feature engineering necessary and the model capacity is not sufficient for handling numerous features. To overcome this issue, deep learning (DL) methods were utilized \cite{goodfellow2016deep}, which extract the valuable features automatically and do not suffer from the curse of dimensionality.

Mapping the CBM data to the RUL of a system is a multivariate time series regression task. Deep learning architectures like convolutional neural networks (CNN) \cite{kim2021multitask} and recurrent neural networks (RNN) \cite{malhi2011prognosis} can efficiently capture temporal and spatial dependencies of condition monitoring (CM) data. CNN methods use receptive fields and multi-dimensional convolutional to extract the feature maps over the time dimension of the CM data. Due to the limitation of the kernel size, only short-term or mid-term temporal dependencies can be captured. Extracting long-term dependencies entails the extension of the kernel size. RNN-based approaches utilize long short-term memory (LSTM) \cite{zheng2017long} or gated recurrent unit (GRU) \cite{chen2019gated} components to extract features. The sequence data need to pass through each unit, which causes forgetting important information contained by the sequence data.

Transformers \cite{vaswani2017attention} are sequence-based models, which became of major interest for recent years. Due to the flexible architecture, the applications range from natural language processing (NLP) \cite{dong2018speech} to computer vision (CV) \cite{parmar2018image}. It utilizes an attention mechanism to model sequence data and extract features. Without considering the distance of the elements in the sequence, it captures short-term and long-term dependencies. The vanilla Transformer only extracts the time series dependencies, whereas the sensor dimension of CBM data contains different type of degradation information. For capturing degradation information in time series dimension and sensor dimension, Dual Aspect Self-attention based on Transformer (DAST) \cite{zhang2022dual} was proposed.

Due to the flexible structure of Transformer models, this architecture makes few assumptions about the structural bias. Especially on small dataset, it leads to poor generalization. Comparing RUL prediction datasets \cite{saxena2008damage} to other datasets \cite{asano21pass} used to train Transformer models, the dataset size tends to be a lot smaller. To overcome this issue, we propose Sparse Low-ranked self-Attention Transformer (SLAT). It has an encoder-decoder architecture, whereas the encoder consists of two parallel blocks to simultaneously make attention to time steps as well as sensors. The generated feature maps of the two encoder blocks are fused and fed into the decoder. To increase the generalization ability on small datasets, SLAT induces structural bias using sparsity in the attention matrix and a low-rank parametrization to the encoder. The main contributions are as follows:
\begin{enumerate}
    \item We introduce critical degradation scenarios in optical fiber amplifier and the corresponding data acquisition setup.
    \item A novel RUL prediction method based on the Transformer architecture is proposed, which utilizes self-attention and dual aspect feature extraction combined with sparsity and low-rank parametrization.
\end{enumerate}

The paper is organized in the following way. At first, the methodology is described in detail. Second, critical components of optical amplifier and their degradation behavior are introduced. Third, the performance increase of the proposed method is shown and finally, a summary of this paper will be given.

\section{METHODOLOGY}
\subsection{PROBLEM DESCRIPTION}
The CBM data of a system at runtime are used for RUL prediction, in this case for an optical fiber amplifier. The data is recorded at defined inspection intervals and has the form $X_{t}\in \mathbb{R}^{k}$ with $t=\left(1,\ldots,T\right)$, where $T$ describes the number of inspection intervals and $k$ the number of sensors. The forecast of the RUL is defined as follows:
\begin{equation}\label{eq:Prog:RULMap}
    y_{t}=f\left(X_{t}\right),
\end{equation}
where $X_{t}$ is the CBM data recorded at runtime, the mapping function $f$ is an arbitrary model, and $y_{t}$ is the predicted RUL. Therefore, the prediction maps the input data matrix $X_{t}$ to a scalar value $y_{t}$. The signal from sensors are relevant for forecasting the RUL. The original approach of the vanilla Transformer only extracts the temporal dependence of the input data. DAST proposed an enhanced architecture that also includes the sensor dependency, as these two dependencies are different in nature and contain information about the degradation of the system in different forms. Basically, Transformer have a flexible structure, as they make few assumptions about the structural bias compared to other models like DCNN \cite{li2018remaining} and BiLSTM \cite{wang2018remaining}. Poor generalization is a result of short sequence lengths and small datasets. Structural bias can be induced by using a sparse attention matrix, such as:
\begin{equation}\label{eq:Prog:SparseAtt}
    \widehat{\mathrm{Attention}}_{ij}=\left\{\begin{array}{l}q_{i}k_{j}^{T}\;\; \mathrm{if\; \textit{i}\; attends\; to\; \textit{j},}\\
    -\infty\;\; \mathrm{if\; \textit{i}\; does\; not\; attend\; to\; \textit{j},}\end{array}\right. .
\end{equation}
This limits the query-key pairs and reduces computation complexity. The $-\infty$ element is not stored in the memory, what results in sparse connections. This mechanism is defined as position-based sparse attention, which uses sparse attention patterns \cite{lin2022survey}. It counteracts the deterioration of modeling long-range dependencies by inserting global nodes. The data also contain local dependencies, which are modeled using a band attention pattern. A combination of both patterns is used for SLAT. Furthermore, the sequence length of CBM data is limited, whereas Transformer were originally designed to handle long sequences. By applying them to short sequence lengths, they tend to overfit. A limitation of the model dimension $D_{model}$ reduces the effect of over-parametrization \cite{guo2019low}. With the combination of sparsity and low-rank parametrization of the attention mechanism, the transformer model SLAT is obtained.
\begin{figure*}[ht!]
    \centerline{\includegraphics[width=0.8\linewidth]{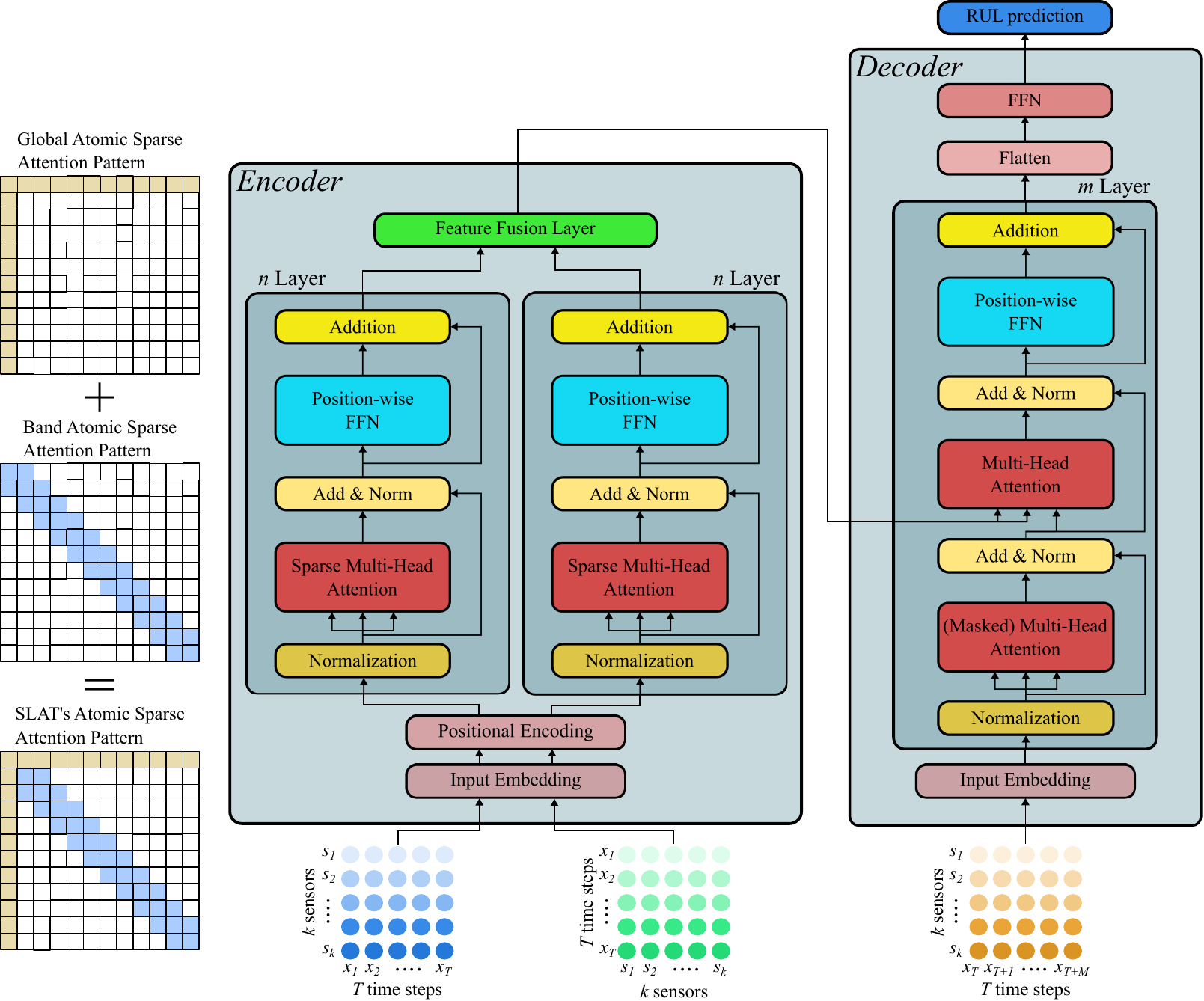}}
    \caption{SLAT Architecture}
    \label{SLAT}
\end{figure*}

The architecture of SLAT is shown in Fig.~\ref{SLAT}. The atomic sparse attention patterns are shown in the left section. A global atomic sparse attention pattern is combined with a band atomic sparse attention pattern that results in SLAT's atomic sparse attention pattern. In the middle section, the encoder is shown. It contains two parallel blocks, which extract the temporal dependencies as well as the sensor dependencies. To achieve that, the input data are provided in the original form and the transposed one. Both are embedded and to add positional information a positional encoding layer is used. Downstream of the encoder blocks, the generated feature maps are fused to build the actual feature map of the encoder.

The decoder has a similar structure as the encoder, but there is no parallelization. Additionally, it utilizes two attention mechanisms. The first attention mechanism is a self-attention, whereas the second is cross-attention with the output of the first attention layer and the feature map of the encoder. Downstream the decoder, a fully connected layer is used as regression component to map the feature map of the decoder to the RUL. In the following, the detailed mode of operation for the encoder and decoder will be discussed.

\subsection{ENCODER-DECODER ARCHITECTURE}
The vanilla Transformer, as well as SLAT, use an encoder-decoder architecture, where each encoder and decoder consists of $n$ and $m$ identical blocks respectively. As the extraction of the feature maps is done by the attention mechanism, those are the most important components in the blocks. The encoder part consists of two parallel blocks, one for feature extraction of the inspection intervals and one for the sensors. By transposing the input matrix of the original data, the input matrix for feature extraction of the sensors is generated. The following explanations of feature extraction process include both encoder blocks. As Transformer require sequential input data, the input embedding is used to map the input data to a vector of dimension $D_{model}$. As the Transformer makes no assumptions about structural bias like DCNN and RNN, positional information must be added to the embedding vector. The positional encoding generates structural information and can be applied in various forms. In this paper the original approach presented in \cite{vaswani2017attention} is used with:
\begin{equation}\label{eq:Prog:PosEncEven}
    PE_{\left(t,2k\right)}=\sin \left(t/10000^{2k/D_{model}}\right),
\end{equation}
\begin{equation}\label{eq:Prog:PosEncOdd}
    PE_{\left(t,2k+1\right)}=\cos \left(t/10000^{2k+1/D_{model}}\right),
\end{equation}
where $t$ and $k$ represent the inspection intervals and sensors respectively. The CBM data matrix has the form $X=\{X_{1},\ldots,X_{T}\}\in\mathbb{R}^{d_{k}\times T}$, where $d_{k}$ is the sensor dimension and $T$ is the number of inspection intervals. For feature extraction of the inspection intervals $X$ is used and for feature extraction of the sensor $X^{T}$ is used. Nowadays two variants of implementing layer normalization exist, \textit{post-LN} \cite{xiong2020layer} and \textit{pre-LN} \cite{wang2019r}, whereas SLAT will use \textit{pre-LN}:
\begin{equation}\label{eq:Prog:NormMH}
    X_{E}=\mathrm{LayerNorm}\left(X_{E}^{pos}\right),
\end{equation}
Downstream the first normalization layer, the multi-head attention (MHA) block follows. It uses a query-key-value model, which results in self-attention if the data matrix is used for query, key, and value respectively:
\begin{equation}\label{eq:Prog:QKV}
    Q_{E}=X_{E}W_{E}^{Q},\;K_{E}=X_{E}W_{E}^{K},\;V_{E}=X_{E}W_{E}^{V},
\end{equation}
where $X_{E}$ is the resulting matrix after the first normalization layer. By multiplying $X_{E}$ with learnable weight matrices, the necessary matrices for $Q$, $K$, and $V$ are formed. With the use of $D_{model}$ as the input dimension, $Q_{E},K_{E},V_{E}\in\mathbb{R}^{d_{k}\times D_{model}}$ follows. Low-rank parametrization is induced by limiting $D_{model}$. The attention of SLAT is created as follows:
\begin{equation}\label{eq:Prog:Att}
    \mathrm{Attention}_{E}=\mathrm{softmax}_{E}\left(\frac{Q_{E}K_{E}^{T}}{\sqrt{D_{model}}}\right)V_{E},
\end{equation}
with $A=\mathrm{softmax}_{E}\left(Q_{E}K_{E}^{T}/\sqrt{D_{model}}\right)$ as \textit{attention matrix}. The \textit{softmax} is applied in a row-wise manner and scaling the attention matrix with $\sqrt{D_{model}}$ alleviates the problem of gradient vanishing. The sparsity of SLAT is implemented here by applying the given attention pattern to the attention matrix. Furthermore, Transformer use multiple attention heads, so that $h$ different projections can be learned, and the performance is improved. The attention for each head is calculated with $head_{i}=\mathrm{Attention}_{E}\left(Q_{E},K_{E},V_{E}\right)_{i}$ and then returned to the original dimensional representation:
\begin{equation}\label{eq:Prog:MH}
    \mathrm{MultiHead}_{E}\left(Q_{E},K_{E},V_{E}\right)=\mathrm{Concat}_{E}\left(\{head_{i}\}_{i=1}^{h}\right)
\end{equation}
Residual connections around each block are used to create deep networks, which are also subject to normalization to avoid gradient problems.
\begin{equation}\label{eq:Prog:NormMid}
    H'=\mathrm{LayerNorm}\left(\mathrm{SelfAttention}\left(X\right)+X\right),
\end{equation}
Additionally, position-wise feed-forward network (FFN) is used with rectified linear unit (ReLU) as activation function, which can be interpreted as convolution with kernel size $1$, whereby additional feature extraction is achieved across the individual projections of MHA. The implementation is as follows:
\begin{equation}\label{eq:Prog:FFN}
    \mathrm{FFN}\left(x\right)=\mathrm{max}\left(0, xW_{1}+b_{1}\right)W_{2}+b_{2}
\end{equation}
By adding another residual connection, the model depth is increased. The aforementioned feature extraction process is performed by two parallel blocks for inspection intervals and sensors respectively. Those blocks can be stacked $n$ times in order to create a deep model. Finally, the created feature maps of the parallel paths are concatenated along one tensor axis and merge into one feature map:
\begin{equation}\label{eq:Prog:Feat}
    F=\mathrm{Concat}\left(F_{E}^{i},F_{E}^{s}\right)W^{F},
\end{equation}
where $F_{E}^{i}$ is the feature map of the inspection intervals, $F_{E}^{s}$ is the feature map of the sensors and $W^{F}\in\mathbb{R}^{\left(d_{k}+T\right)\times D_{model}}$ is a trainable weight matrix of the fusion layer.

The decoder of SLAT is similarly implemented to the encoder. The data matrix of the encoder differs from to the decoder as different inspection intervals are present in the data matrix, which results in $X=\{X_{T},\ldots,X_{T+M}\}$, where $M$ represents the number of current inspection intervals. The choice of the ratio of $T$ and $M$ will be clarified as part of the structural analysis. The data matrix is embedded in the first step, followed by a normalization layer. In the decoder, no positional encoding is used, as the data matrix only represents the most recent inspection interval. In contrast to the encoder, the decoder uses two MHA blocks inside one decoder block. The first one uses a self-attention mechanism with residual connections. The second one, resembles a cross-attention mechanism of the generated feature map of the encoder to the previous feature map of the decoder. This enables a conjuncted analyzation of data from earlier inspection intervals with current inspection intervals. The combined projection of the cross-attention block is used as input for the position-wise FFN. Similar to the encoder, the decoder uses \textit{pre-LN} as implementation type for layer normalization. The decoder blocks can also be stacked $m$ times to create a deep model. Downstream the decoder block a flattening and FFN layer is used for the regressive task of mapping the generated feature map to the RUL.

\section{EXPERIMENTAL DESIGN}
\subsection{DEGENERATION EFFECTS}
Nowadays, there exist many types of optical amplifiers. Quite popular are rare-earth doped fiber amplifier. They are characterized by the choice of the rare earth ions as dopant for the amplification fiber \cite{rapp2021optical}. Due to the sharp spectral energy bands of the lanthanides, they achieve amplification in different frequency ranges. Most of the transmission systems operate in the C-band using erbium-doped fiber amplifier (EDFA). Therefore, we will focus on this type in the following.
\begin{figure}[h]
    \centering
        \includegraphics[width=\linewidth]{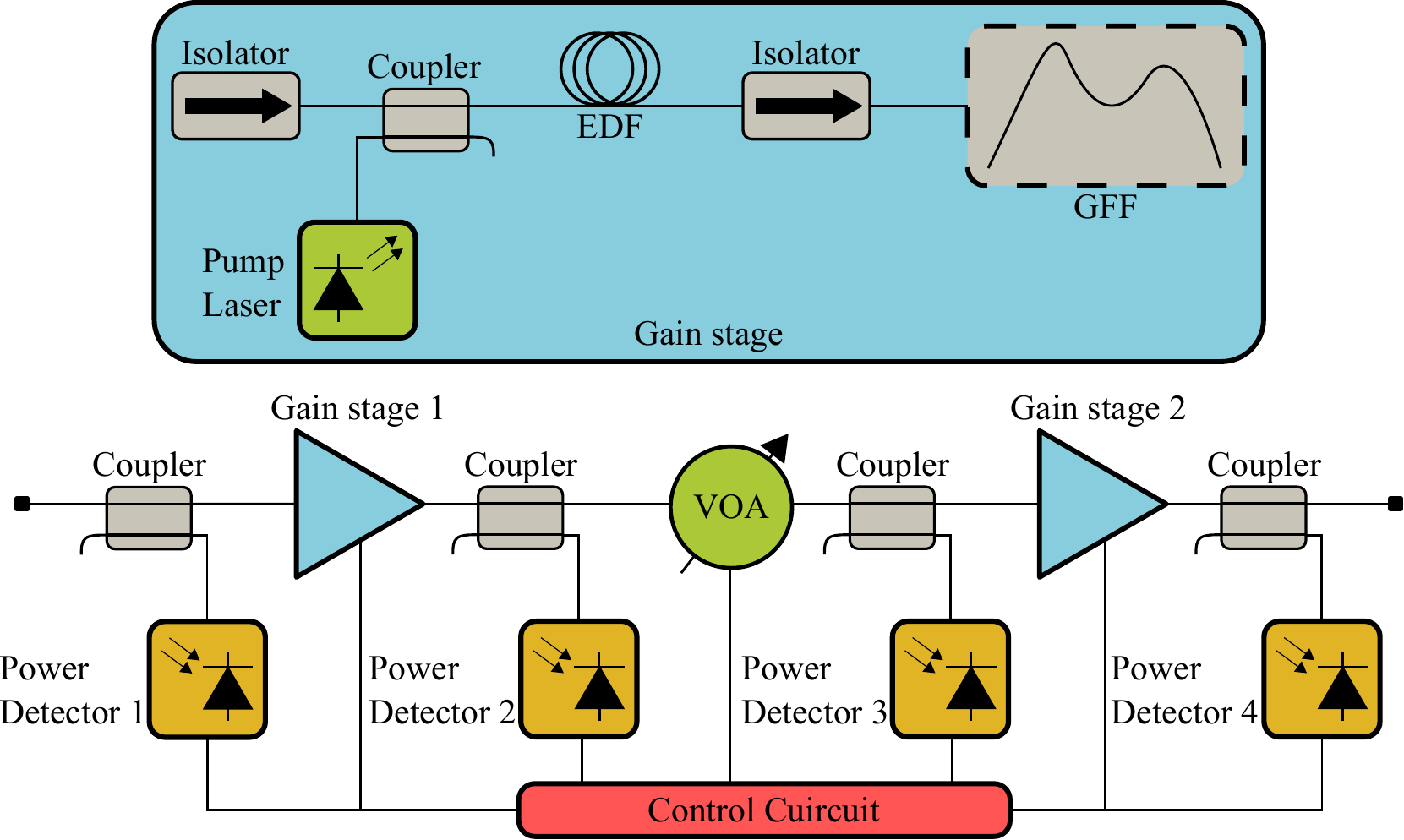}
    \caption{Double-stage EDFA}
    \label{P:fig:EDFA}
\end{figure}
Commercially used optical fiber amplifiers are composed of the components, as shown in Fig.~\ref{P:fig:EDFA}. Typically, those amplifiers have an automatic gain control (AGC) with two independent gain stages. The first stage has two power detectors to measure the total input and output power of the stage. To adjust the gain, a control mechanism adjusts the pump current and thus the pump power, which is necessary for the amplification process in the amplification fiber. As the gain is dependent on the wavelength of the individual signal, a gain-flattening filter (GFF) is used to alleviate this effect. Downstream of the first gain stage, a variable optical attenuator is placed to adjust the total input power of the second stage. This is used for tuning the overall tilt of the system. The second gain stage has the same components as the first, except for the GFF. The system gain is composed of the gain of both stages and an interstage loss created by the variable optical attenuator (VOA), as follows:
\begin{equation}\label{Gain}
    G=G_{1}+G_{2}+LOSS_{INT}.
\end{equation}
The total gain of the system can be adjusted. By utilizing a lookup table, which is created during the design of the amplifier, the independent values for $G_{1}$ and $G_{2}$ are set, and $LOSS_{INT}$ can be calculated from \ref{Gain}.

For applying RUL prediction to optical amplifier in the context of PdM, the most critical components need to be identified. With the use of a critical component selection procedure \cite{tinga2019physical}, utilizing criticality classification, showstopper identification, and focused feasibility study, four groups could be identified:
\subsubsection{Pump Laser}
The first components to be considered are the pump laser, installed in gain stage one and two respectively. Those are the main actuators of the gain control. The task of the pump is to bring the rare earth ions in the amplification fiber to a higher energy level, thus it enables the amplification process. The optical output power of the pump laser is current-controlled. This relation is linear in nature. With degradation of the pump, the required driving current increases to achieve a desired optical output power \cite{6836254} \cite{Bliznyuk2021} \cite{9670373}. The gain control will compensate this effect until the desired power level is not reachable anymore, which will result in a decrease of the actual gain of the specific stage.
\subsubsection{Power Detector}
Power detectors (PD) are the measurement components in the system, that give feedback of the actual values to the control unit. Technically, photodiodes are used as power detectors, whereas degradation results in a decrease of the sensitivity to incident light \cite{xu2013degradation} \cite{huang2016predictive}. This effects the feedback values into the control shift and thus the gain control as the operating conditions change.
\subsubsection{Variable Optical Attentuator}
The VOA is, besides both pump laser, an actuator in the control loop of the optical amplifier. It is responsible for setting the correct value of the interstage loss. This component is controlled using a driving voltage, whereas when not controlled the resulting attenuation is preferably at its lowest value. Degradation results in a lower attenuation for a specific driving voltage compared to BOL \cite{kowalczyk2004polymer} \cite{llobera2006polymeric}. This has no effect onto the first stage, but increases the incoming total power of the second stage and therefore changes the operating point and the system tilt.
\subsubsection{Passive Components}
There are three types of passive components (PC) in the system. The first PCs are optical couplers. They couple incident light of one common port to two other ports with a given splitting ratio. At BOL, the PD are calibrated to the power level in the signal path. Degradation reduces the transmittance of the coupler \cite{bednarek2016influence}. The second PCs are isolators. They suppress the back propagating amplified spontaneous emission (ASE) to enhance the efficiency of the amplifier. In forward direction, they are characterized by a given transmittance and in backward direction through attenuation to ensure the blocking of light. Analogous to optical couplers, degradation effects result in lower transmittance \cite{tomasi2003passive}. The final PC is the GFF, which alleviates the wavelength-dependence of the gain to achieve a flat gain spectrum, by inserting wavelength-dependent loss \cite{lancry2014reliable}. All types of PC degradation effects the behavior of the system, depending on their position.

\subsection{DATA ACQUISITION SETUP}
For data acquisition, a wavelength-division multiplexing (WDM) signal with nine equally distributed channels is used, which is generated by independent tunable lasers in the C-band. A multiplexer combines spectral lines of the lasers. Commercial optical amplifier can operate at a given total input power range from $-35\;\mathrm{dBm}$ to $1\;\mathrm{dBm}$, thus a VOA is used to vary the total input power to the system. To create two selective signal paths, one for the input signal and one for the output signal of the optical fiber amplifier, two optical switches are used. As measurement devices a power meter (PM) and an optical spectrum analyzer (OSA) is used.
\begin{figure}[h]
    \centerline{\includegraphics[width=\linewidth]{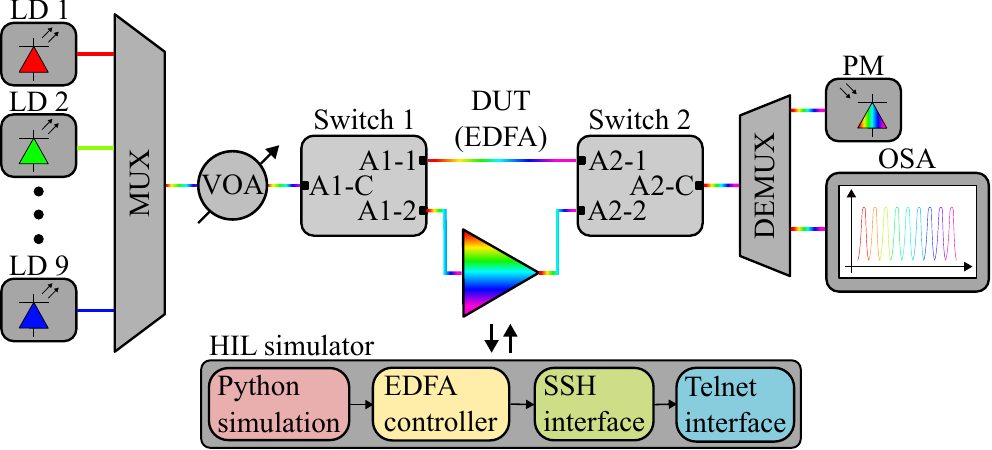}}
    \caption{Data Acquisition Setup}
    \label{ProgHW}
\end{figure}
The described setup is shown in Fig.~\ref{ProgHW}. To induce the degradation effects into the system, a hardware-in-the-loop (HIL) simulator is used. The degradation script controls the optical fiber amplifier through a control engine that communicates with the hardware (HW) through a secure shell (SSH) and a telnet interface. By manipulation of certain registers on the controller of the optical fiber amplifier, each degradation effect can be generated.

\subsection{PARAMETRICAL ANALYSIS}
The aforementioned data acquisition setup was used to record the EDFA-specific optical fiber amplifier system simulation (E-OFASS) dataset. The four groups of degradation effects, created by the individual components, form the four sub-datasets of E-OFASS, whereas \textit{FD1} includes pump degradation, \textit{FD2} includes PD degradation, \textit{FD3} includes VOA degradation, and \textit{FD4} includes PC degradation.
\begin{table}[h]
    \centering
    \caption{E-OFASS}
    \label{tab:P:eofass}
    \begin{tabular}{|l|c|c|c|c|}
        \hline
        \multirow{2}{*}{\textbf{Dataset}} & \multicolumn{4}{c}{\textbf{E-OFASS}}\vline\\
        \cline{2-5} & FD1 & FD2 & FD3 & FD4\\
        \hline
        Operating conditions & 153 & 153 & 153 & 153\\
        Fault modes & 2 & 4 & 1 & 4\\
        Training set size & 92100 & 184100 & 46950 & 184100\\
        Test set size & 27652 & 56069 & 12980 & 55835\\
        \hline
    \end{tabular}
\end{table}
The dataset is shown in Table.~\ref{tab:P:eofass}. Each of the sub-datasets was recorded for the same operating conditions, varying the input power and the gain of the system in range of $19\;\mathrm{dB}$ to $35\;\mathrm{dB}$. The fault modes represent the individual components, which differ in each degradation group and thus leads to various training and test set sizes. The sub-datasets contain CBM data of different inspection intervals. As degradation is evolving over time, in the early inspections the RUL will be limited. Therefore, we will follow previous work and set $RUL_{max}=125$ \cite{zhang2022dual}. This is especially for comparison of the prediction results necessary, as those will differ for different settings of $RUL_{max}$. The splitting ratio of training and test sets is $0.33$, whereas the test set contains independent samples of the training set. Furthermore, the test datasets do not contain complete run-to-failure (RTF) as the RTF is stopped at earlier inspection intervals.

\begin{figure}[h]
    \centering
        \includegraphics[width=\linewidth]{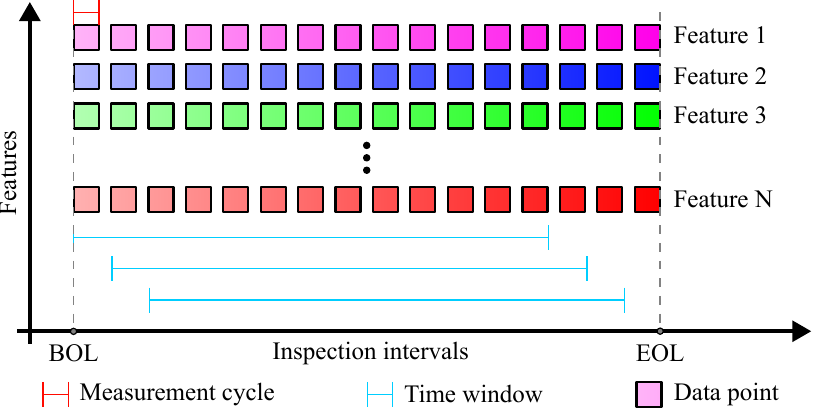}
    \caption{STW technique}
    \label{P:fig:STW}
\end{figure}

For data preprocessing, the data are normalized by applying a min-max scaler to the CBM data $X_{i}=\{X_{1},\ldots,X_{T}\}$, as DL algorithms are sensitive to non-scaled data \cite{ahsan2021effect}:
\begin{equation}\label{eq:Prog:minmax}
    \widehat{X_{i}}=\frac{X_{i}-\min\left(X_{i}\right)}{\max\left(X_{i}\right)-\min\left(X_{i}\right)},
\end{equation}
where $\widehat{X_{i}}$ denotes the normalized CBM data. To take advantage from multivariate time series, a sliding time window (STW) technique is used. A fixed length for the time window $T_{w}$ is defined. The technique is illustrated in Fig.~\ref{P:fig:STW}.
In addition, two statistical features are added to the CBM data to provide contextual information about the time series to the DL model \cite{chen2020machine}. As first feature, the statistical mean is used:
\begin{equation}\label{eq:Prog:Mean}
    \overline{X_{i}}=\frac{1}{T}\sum_{n = 1}^{T}X_{i}^{n}.
\end{equation}
The second statistical feature uses the regression coefficients of a linear regression, as follows:
\begin{equation}\label{eq:Prog:LinReg}
    Y_{i}=a*X_{i}+b,
\end{equation}
where $Y_{i}$ defines the following time step of the input data $X_{i}$ provided to the linear regression model. The fitted coefficients $a$ and $b$ as well as the mean $\overline{X_{i}}$ are appended to the CBM data. For performance evaluation of the DL models, the root-mean-square-error (RMSE) is used.

\section{RESULTS AND ANALYSIS}
\subsection{STRUCTURAL ANALYSIS}
SLAT contains various hyperparameters that have significant impact on the performance of the model. In order to determine the best configuration, Bayesian optimization is performed.
\begin{table}[h]
    \centering
    \caption{Structural Parameters of SLAT}
    \label{tab:StructParam}
    \begin{tabular}{|c|c|c|}
        \hline
        \textbf{Components} & \textbf{Layers} & \textbf{Parameters}\\
        \hline \multirow{2}{*}{Input embedding} & \multirow{2}{*}{Fully connected layer} & Hidden units: $64$\\
        & & Activation: Linear\\
        \hline \multirow{4}{*}{Encoder} & \multirow{2}{*}{Sensor encoder layer} & Blocks: $N = 4$\\
        & & Heads: $H=8$\\
        \cline{2-3}
        & \multirow{2}{*}{Time step encoder layer} & Blocks: $N = 4$\\
        & & Heads: $H=8$\\
        \hline \multirow{2}{*}{Decoder} & \multirow{2}{*}{Decoder layer} & Blocks: $N=2$\\
        & & Heads: $H=8$\\
        \hline \multirow{4}{*}{Output} & \multirow{2}{*}{Fully connected layer} & Hidden units: $64$\\
        & & Activation: ReLU\\
        \cline{2-3}
        & \multirow{2}{*}{Output layer} & Hidden units: $1$\\
        & & Activation: Linear\\
        \hline
    \end{tabular}
\end{table}
Table.~\ref{tab:StructParam} lists the resulting structural parameters. Additionally, the window length of the STW procedure is set to $40$ for all sub-datasets. For training, the Adam optimizer is applied with $\beta_{1}=0.9$, $\beta_{2}=0.98$, and $\epsilon=1e-9$. For Transformer models it is necessary to implement a learning rate scheduler to provide a warm-up phase for the algorithm and to ensure a stable learning behavior. We will follow the common approach \cite{vaswani2017attention} with:
\begin{equation}\label{eq:Sched}
    lr=D_{model}^{-0.5}\cdot \min\left(sn^{-0.5},sn\cdot ws^{-1.5}\right),
\end{equation}
where $lr$ is the learning rate, $sn$ the step number, and $ws$ the warm-up steps, which is set to $ws=4000$. The total number of training epoch is set to $300$. The RMSE is used as loss function for training and as objective for regularization using the early stopping technique. The experiments are carried out on a Windows $11$ workstation with $160$ GB RAM and an NVIDIA RTX A5000 graphics card. To alleviate stochastic influence of the random weight initialization, the number of runs is set to $25$. The source code is available at \url{https://github.com/DomSchResearch/SLAT} and the trained model at \url{https://huggingface.co/dschneider96/SLAT}.

\subsection{COMPARISON TO OTHER MODELS}
The performance of SLAT is compared to other state-of-the-art methods. Therefore, the widely adopted commercial modular aero propulsion system simulation (C-MAPSS) \cite{saxena2008damage} dataset is used. It contains, equally to E-OFASS, four sub-datasets.
\begin{table}[h]
    \centering
    \caption{C-MAPSS}
    \label{tab:P:cmapss}
    \begin{tabular}{|l|c|c|c|c|}
        \hline
        \multirow{2}{*}{\textbf{Dataset}} & \multicolumn{4}{c}{\textbf{C-MAPSS}}\vline\\
        \cline{2-5} & FD001 & FD002 & FD003 & FD004\\
        \hline
        Training engines & 100 & 260 & 100 & 249\\
        Testing engines & 100 & 259 & 100 & 248\\
        Operating conditions & 1 & 6 & 1 & 6\\
        Fault modes & 1 & 1 & 2 & 2\\
        Training set size & 20631 & 53759 & 61249 & 45918\\
        Test set size & 100 & 259 & 100 & 248\\
        \hline
    \end{tabular}
\end{table}
As shown in Table.~\ref{tab:P:cmapss} each sub-dataset contains different numbers of engines and operating conditions, as well as fault modes. In the previous work, \textit{FD001} and \textit{FD003} were trained with a time window size of $40$ and \textit{FD002} and \textit{FD004} with a time window size of $60$, due to their complexity. Data preparation and preprocessing are carried out as described in the references.

Four comparable categories were identified: 1) pure RNN/CNN based methods \cite{liu2020remaining} \cite{li2018remaining} \cite{wang2018remaining}, 2) combinations of RNN/CNN with attention \cite{song2020distributed} \cite{zeng2021deep}, 3) methods based on health indicators \cite{kong2019convolution}, and 4) transformer based methods \cite{zhang2022dual}. At first, SLAT is compared against the baseline methods, shown in Table.~\ref{tab:AD:cmapssResults}. It is noticeable that SLAT outperforms all methods on each dataset, which means that a more precise RUL prediction is done. Especially the performance improvement of SLAT against the baseline methods tends to be better for the more complex sub-datasets \textit{FD002} and \textit{FD004}. This leads to the conclusion that SLAT has a better generalization ability and can handle complex multivariate time series data better than state-of-the-art methods.
\begin{table*}[t]
    \centering
    \caption{RMSE Performance evaluation on C-MAPSS dataset}
    \label{tab:AD:cmapssResults}
    \begin{tabular}{|l|c|c|c|c|c|c|c|c|}
        \hline
        \textbf{Dataset} & \textbf{BiLSTM} \cite{wang2018remaining} & \textbf{DCNN} \cite{li2018remaining} & \textbf{Kong et al.} \cite{kong2019convolution} & \textbf{DATCN} \cite{song2020distributed} & \textbf{DARNN} \cite{zeng2021deep} & \textbf{AGCNN} \cite{liu2020remaining} & \textbf{DAST} \cite{zhang2022dual} & \textbf{SLAT}\\
        \hline
        FD001 & 13.65 & 12.61 & 16.13 & 11.78 & 12.04 & 12.41 & 11.43 & \textit{11.26}\\
        FD002 & 23.18 & 22.36 & 20.46 & 16.95 & 19.24 & 19.43 & 15.25 & \textit{14.23}\\
        FD003 & 13.74 & 12.64 & 17.12 & 11.56 & 10.18 & 13.39 & 11.32 & \textit{11.24}\\
        FD004 & 24.86 & 23.31 & 23.26 & 18.23 & 18.02 & 21.50 & 18.36 & \textit{16.81}\\
        \hline
        Average & 18.85 & 17.73 & 19.24 & 14.63 & 14.87 & 16.68 & 14.09 & \textit{13.39}\\
        \hline
    \end{tabular}
\end{table*}
Then, the prediction accuracy of SLAT will be visually analyzed. Therefore, an engine from each sub-dataset is selected and the predicted RUL is compared to the actual RUL in an RTF. For this experiment, the BiLSTM \cite{wang2018remaining} and DCNN \cite{li2018remaining} methods are used, as well as the transformer-based method DAST \cite{zhang2022dual}.
\begin{figure}[t]
    \centering
        \includegraphics[width=\linewidth]{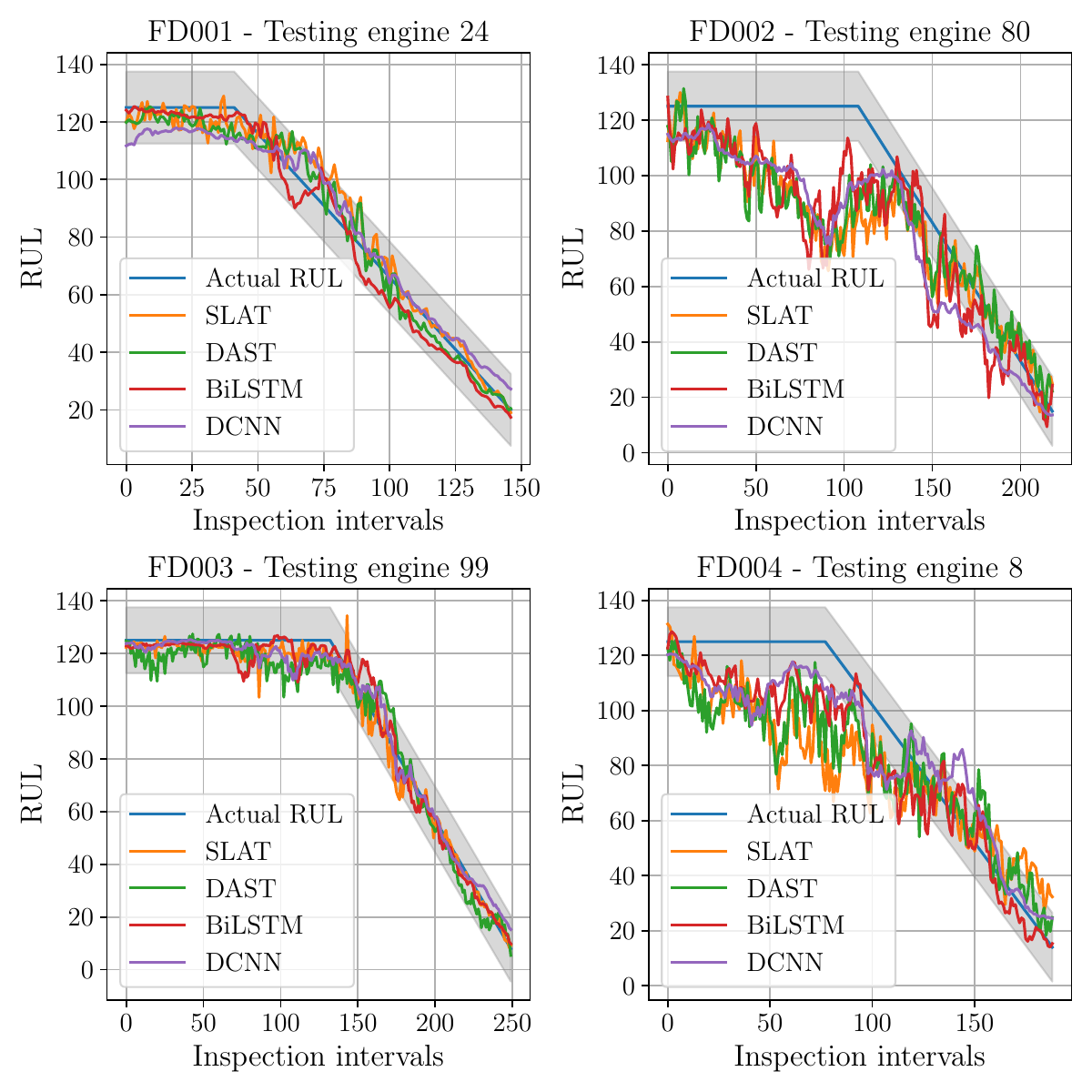}
    \caption{C-MAPSS RTF}
    \label{P:fig:cmapss_RTF}
\end{figure}
In Fig.~\ref{P:fig:cmapss_RTF} the actual RUL and a confidence interval (CI) with $[-0.1\cdot RUL_{max},0.1\cdot RUL_{max}]$ is shown. The predicted RUL of SLAT follows the trajectory of the actual RUL better than the reference architectures. This shows, that the proposed method is able to capture the information of degradation.

\subsection{FORECASTING RESULTS}
SLAT has shown superior performance compared with state-of-the-art methods on the reference dataset C-MAPSS. In the following, SLAT is compared against BiLSTM, DCNN and DAST on the E-OFASS dataset, shown in Table.~\ref{tab:AD:ofaResults}.
\begin{table}[t]
    \centering
    \caption{RMSE Performance evaluation on E-OFASS dataset}
    \label{tab:AD:ofaResults}
    \begin{tabular}{|l|c|c|c|c|}
        \hline
        \textbf{Dataset} & \textbf{BiLSTM} \cite{wang2018remaining} & \textbf{DCNN} \cite{li2018remaining} & \textbf{DAST} \cite{zhang2022dual} & \textbf{SLAT}\\
        \hline
        FD1 & 10.18 & 9.72 & 9.66 & \textit{8.93}\\
        FD2 & 9.03 & 8.75 & 8.28 & \textit{7.67}\\
        FD3 & 3.73 & 2.70 & 1.58 & \textit{1.34}\\
        FD4 & 10.81 & 10.20 & 9.99 & \textit{8.29}\\
        \hline
        Average & 7.72 & 7.85 & 7.5 & \textit{6.56}\\
        \hline
    \end{tabular}
\end{table}
\begin{figure}[b!]
    \centering
        \includegraphics[width=\linewidth]{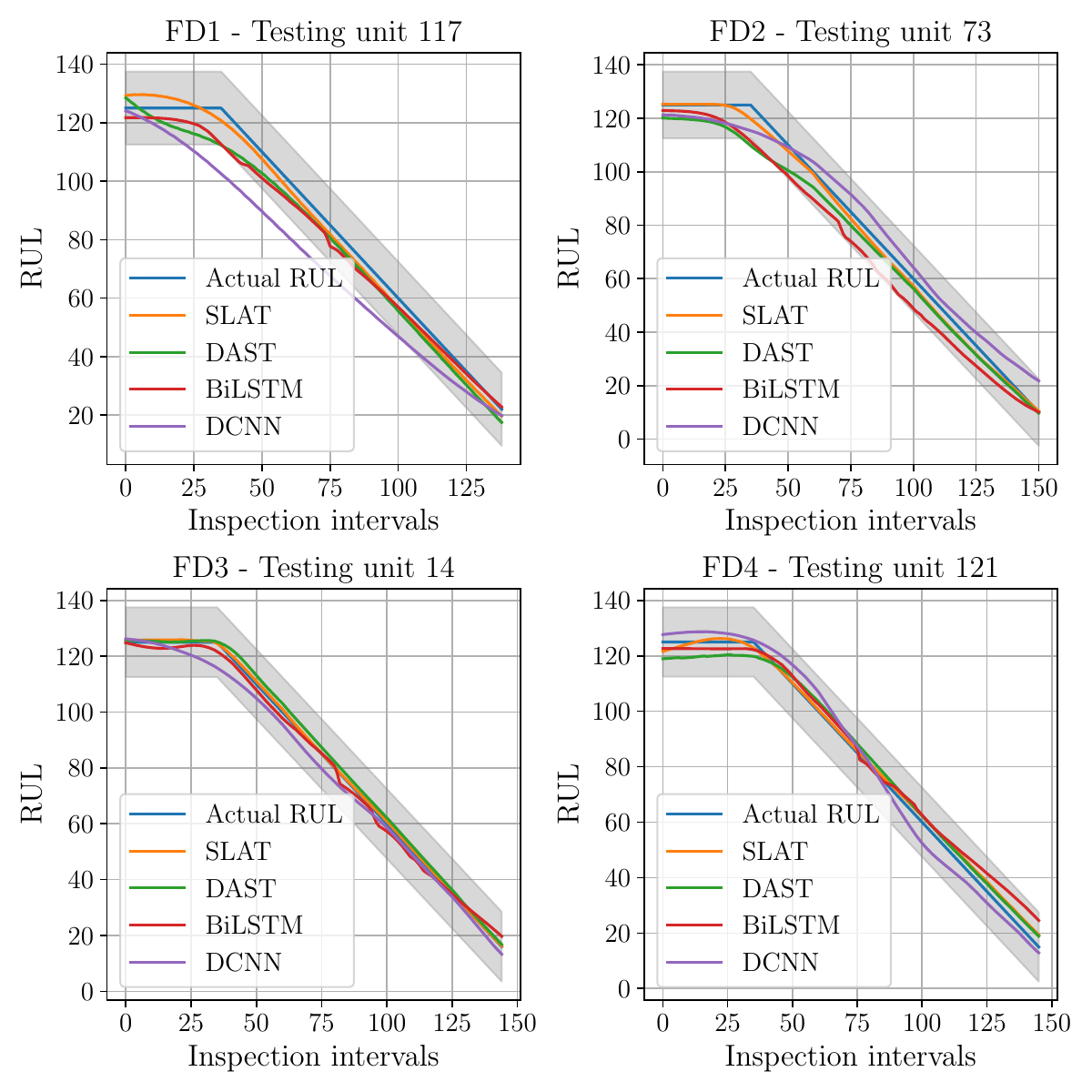}
    \caption{E-OFASS RTF}
    \label{P:fig:eOFASS_RTF}
\end{figure}
First, it is noticeable that all methods achieve a better RMSE score on E-OFASS than C-MAPSS, which leads to the conclusion, that the containing degradation information is easier to learn. Furthermore, SLAT is outperforming the other models for each sub-dataset. SLAT reduces the RMSE for \textit{FD1}, \textit{FD2}, \textit{FD3}, and \textit{FD4} by $7.6\%$, $7.4\%$, $15.2\%$, and $17.0\%$ respectively. Especially the performance increase on the smallest dataset, supports the statement that SLAT has an increased generalization capability.
For a visual analyzation of the RTF on the E-OFASS dataset, different testing units from the sub-dataset are randomly chosen. As in the RMSE comparison, SLAT is compared against BiLSTM, DCNN, and DAST. The results are shown in Fig.~\ref{P:fig:eOFASS_RTF}.

It can be seen, that our proposed method follows the actual RUL of the system similarly. The trajectory is clearly better than those of the compared methods. Additionally, when compared against the C-MAPSS dataset, fewer ripples can be seen, which supports the previous assumption, that the E-OFASS dataset is less complex compared to the C-MAPSS dataset.

\begin{figure}[t]
    \centering
        \includegraphics[width=\linewidth]{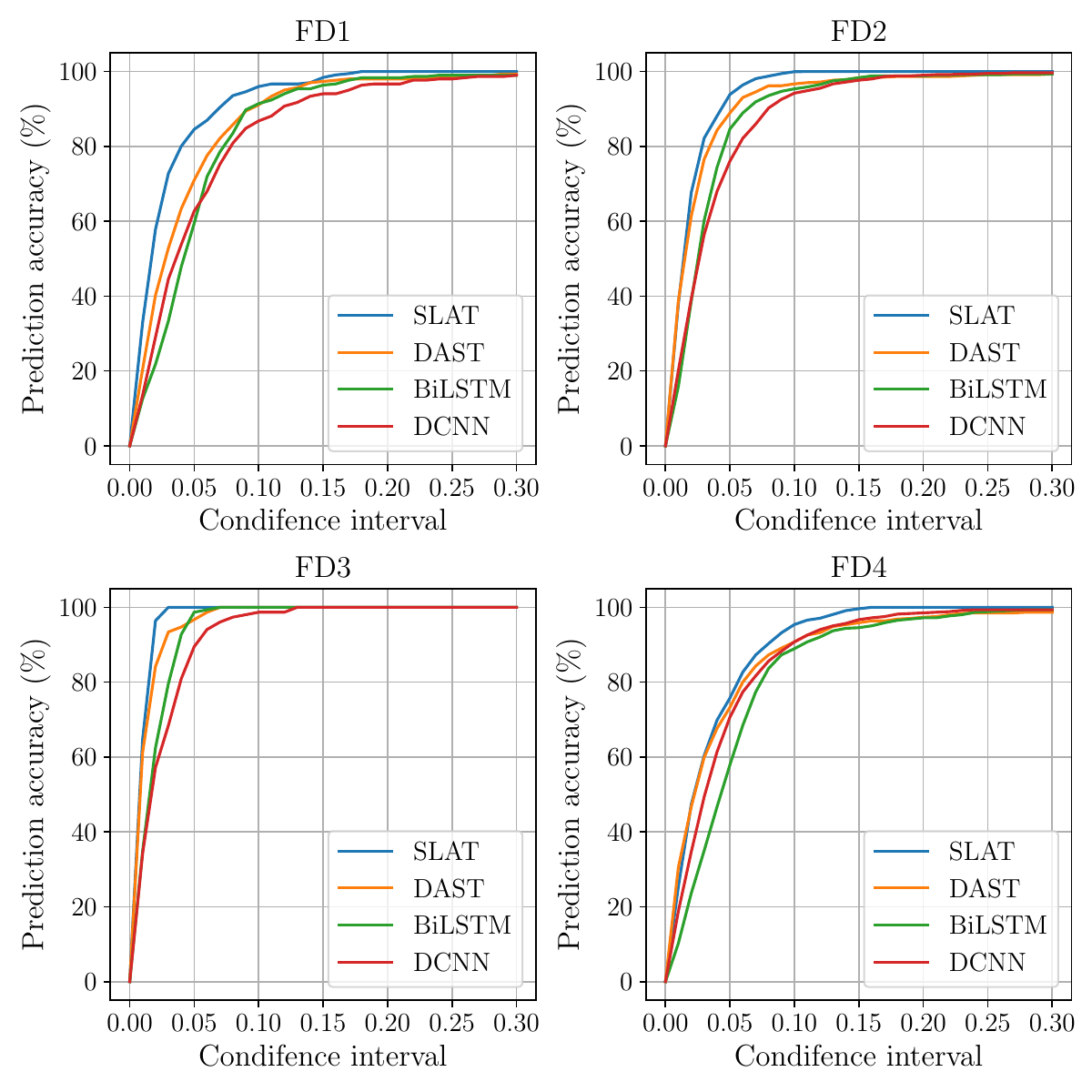}
    \caption{E-OFASS CI}
    \label{P:fig:eOFASS_CI}
\end{figure}

Previously, the comparison of SLAT to other models was carried out using the RMSE as metric or visual analyzation with RTF. For measuring the scattering of the predicted RUL, a CI scoring will be used. This will give an overview how many predictions of the RUL will be in a certain CI. For this purpose, SLAT will be compared to the aforementioned methods. The CI ranges from $[-0.00\cdot RUL_{max}, 0.00\cdot RUL_{max}]$ to $[-0.30\cdot RUL_{max}, 0.30\cdot RUL_{max}]$. Fig.~\ref{P:fig:eOFASS_CI} shows the results of the CI scoring.
It can be seen, that the increase of the prediction accuracy is the steepest for SLAT in all sub-datasets, which means less scattering of the predicted RUL than for the other models For the less complex sub-datasets the increase is higher than for the more complex ones. The $100\;\%$ prediction accuracy is reached by SLAT for \textit{FD1}, \textit{FD2}, \textit{FD3}, and \textit{FD4} in a CI of $0.17$, $0.09$, $0.03$, and $0.14$, respectively. These results imply, that the predicted RUL of SLAT is deviating fewer from the actual RUL than for other models. Therefore, the prediction can be considered more robust than for the reference methods.
\begin{table}[h]
    \centering
    \caption{Train and test times of SLAT}
    \label{tab:Prog:TTT}
    \begin{tabular}{|c|c|c|}
        \hline
        \textbf{Dataset} & \textbf{Train dataset} & \textbf{Test dataset}\\
        \hline
        FD1 & 53 min 28 s & 0.8 s\\
        FD2 & 102 min 54 s & 1.1 s\\
        FD3 & 28 min 43 s & 0.7 s\\
        FD4 & 102 min 55 s & 1.1 s\\
        \hline
        Average & 72 min 0 s & 0.9 s\\
        \hline
    \end{tabular}
\end{table}
In addition, prognostic methods need to be able to forecast the RUL in real-time. In Table.~\ref{tab:Prog:TTT} the training and test times per run for the sub-datasets of E-OFASS are shown. The training times scales linear with the set size. In comparison, the testing times are significantly lower than the training times. As the testing routine resembles inference of the model, no adaptions to the weights are done. Optical fiber amplifier are decentralized systems, embedded in an optical network architecture. With inference times ranging from $0.7\;\mathrm{s}$ to $1.1\;\mathrm{s}$, SLAT is suitable for RUL prediction in real-time.

\section{CONCLUSION}
In this paper, we propose a novel deep learning model for remaining useful lifetime (RUL) prediction of optical fiber amplifiers, exemplified on EDFA. Sparse Low-ranked self-attention Transformer (SLAT) has an encoder-decoder architecture, that uses dual aspect mechanism to extract feature information of time steps and sensors. Enhancing the attention mechanism in the encoder by sparsity and low-ranked parametrization, enables the model to increase the generalization capability. SLAT learns the information included in time steps and sensors automatically, which is to improve maintenance safety and reliability. We identified the critical components in optical fiber amplifier and introduced their degradation behavior. By inducing degradation into an EDFA the E-OFASS dataset was generated. The experimental results on E-OFASS and C-MAPSS dataset prove the superior RUL prediction capability of our method. In the future, we would like to apply SLAT to other core components in optical transmission networks, which entails critical component analysis and degradation behavior analysis. Especially providing feedback through the proposed method into high level network orchestration is of interest for automatic network reconfiguration and reduction of network downtime and increase in network availability.

\section*{ACKNOWLEDGMENT}
D. Schneider thanks Mohammad Taleghani for his support in providing necessary information about the control architecture of the optical fiber amplifiers to induce degradation scenarios and thus generate the E-OFASS dataset.

\bibliographystyle{IEEEtran}
\bibliography{IEEEabrv, references.bib}

\end{document}